\newcommand{\remove}[1] {}%{\textcolor{blue}{#1}}
\newcommand{\ignore}[1]{}

\documentclass[10pt,twocolumn,letterpaper]{article}

\usepackage{cvpr}
\usepackage{times}
\usepackage{graphicx}
 \usepackage{amsmath, amsthm, amssymb, dsfont}
\usepackage{color, colortbl}
\definecolor{myColor}{rgb}{0.95,0.95,0.95}

\usepackage{epsfig}
\usepackage{floatrow}
\usepackage{array}
\usepackage{slashbox}
\usepackage{multirow}
\usepackage{caption}
\usepackage{array,arydshln}

\usepackage{enumitem}

\usepackage{algorithm}
\usepackage{algorithmic}

\usepackage[numbers,sort&compress,square,comma]{natbib}
\usepackage[pagebackref=false,breaklinks=true,letterpaper=true,colorlinks,bookmarks=false]{hyperref}
\newcommand{\oldtext}[1]{} %{\textcolor{orange}{#1}}

\DeclareMathAlphabet{\mathpzc}{T1}{pzc}{m}{n}

\cvprfinalcopy

\def\cvprPaperID{829} % *** Enter the CVPR Paper ID here

\ifcvprfinal\fi

\usepackage{authblk}

\begin{document}
\newtheorem{mydef}{Definition}
\newtheorem{observation}{Observation}

\title{Joint calibration of Ensemble of Exemplar SVMs \vspace{-2mm}}

\author{Davide Modolo\textsuperscript{1}, Alexander Vezhnevets\textsuperscript{1}, Olga Russakovsky\textsuperscript{2}, Vittorio Ferrari\textsuperscript{1} \vspace{-2mm}\\
\it \textsuperscript{1}University of Edinburgh \hspace{0.9cm}\textsuperscript{2}Stanford University  \vspace{-5mm}}
%\author[1]{Davide Modolo}
%\author[1]{Alexander Vezhnevets}
%\author[2]{Olga Russakovsky}
%\author[1]{Vittorio Ferrari}
%\affil[1]{\it University of Edinburgh, Edinburgh, UK} 
%\affil[2]{\it Stanford University, Stanford, CA, USA}
%%\affil[2]{Google DeepMind, London, UK}

\maketitle
\thispagestyle{empty}
% 
%%%%%%%% ABSTRACT
\begin{abstract}
\vspace{-4mm}
We present a method for calibrating the Ensemble of Exemplar SVMs model. Unlike the standard approach, which calibrates each SVM independently, our method optimizes their joint performance as an ensemble. We formulate joint calibration as a constrained optimization problem and devise an efficient optimization algorithm to find its global optimum.
The algorithm dynamically discards parts of the solution space that cannot contain the optimum early on, making the optimization computationally feasible.
%To scale the method to large datasets, we show how to relax global optimality and attain a good approximate solution.
%Experiments on the ILSVRC 2014 and PASCAL VOC 2007 datasets show that
We experiment with EE-SVM trained on state-of-the-art CNN descriptors. Results on the ILSVRC 2014 and PASCAL VOC 2007 datasets show that
(i) our joint calibration procedure outperforms independent calibration on the task of classifying windows as belonging to an object class or not; and
(ii) this improved window classifier leads to better performance on the object detection task.
\end{abstract}

%%%%%%%% BODY TEXT
% ------------------------------------------------------- INTRODUCTION
\vspace{-4mm}
\section{Introduction} \label{sec:intro}
\vspace{-2mm}

The Ensemble of Exemplar SVMs~\cite{MalisiewiczICCV11} (EE-SVM) is a powerful non-parametric approach to object detection. It is widely used~\cite{Shrivastava2011sgrapha, singh12eccv, aytar12bmvc, endres13cvpr, dong13cvpr, tighe13cvpr, song14eccv, juneja13cvpr, gronat13cvpr, vezhnevets14cvpr, aubry2014cvpr} because it explicitly associates a training example to each object it detects in a test image. This enables transferring meta-data such as segmentation masks~\cite{MalisiewiczICCV11,tighe13cvpr}, 3D models~\cite{MalisiewiczICCV11}, viewpoints~\cite{aubry2014cvpr}, GPS locations~ \cite{gronat13cvpr} and part-level regularization~\cite{aytar12bmvc}.
Furthermore, EE-SVM can also be used for discovering objects parts~\cite{endres13cvpr, singh12eccv}, scene classification \cite{singh12eccv,juneja13cvpr}, object classification~\cite{dong13cvpr}, image parsing \cite{tighe13cvpr}, image matching~\cite{Shrivastava2011sgrapha}, automatic image annotation \cite{vezhnevets14cvpr} and 3D object detection \cite{song14eccv}. 

An EE-SVM is a large collection of linear SVM classifiers,
each trained from one positive example and many negative ones (an E-SVM).
At test time each window is scored by all E-SVMs, and the highest score is assigned to the window.
Because of this max operation, it is necessary to calibrate the E-SVMs to make their scores comparable.
A common procedure is to calibrate each SVM independently, by fitting a logistic sigmoid to its output on a validation set~\cite{MalisiewiczICCV11}.
Such independent calibration, however, does not take into account that the final score is the max over many E-SVMs. 
Moreover, calibrating one E-SVM in isolation requires choosing which positive training samples it should score high and which ones it can afford to score low. Such a prior association of positive training samples to E-SVMs is arbitrary, as there is no predefined notion of how much and in which way a particular E-SVM should generalize. What truly matters is the interplay between all E-SVMs through the max operation.

%\todo{rewritten the riff to be correct. now we dance around thresholds. Vitto's version in comments}
In this paper we present a {\em joint} calibration procedure that takes into account the max operation.
We calibrate all E-SVMs at the same time by optimizing their joint performance {\em after} the max.
Our method finds a threshold for each E-SVM, so that
(i) all positive windows are scored positively by at least one E-SVM, and
(ii) the number of negative windows scored positively by any E-SVM is minimized.
The first criterion ensures that there are no positive windows scored negatively after the max, while the second criterion minimizes the number of false positives.

We formalize these two criteria in a well-defined constrained optimization problem. The first requirement is formalised in its constraints, while the second comes in as a loss function to be minimized. Each threshold defines which training samples the respective E-SVM is scoring positively. By lowering a threshold we cover more positives and thereby satisfy more constraints, but we also include more negatives and therefore suffer a greater loss. Any positive sample can be potentially covered by any E-SVM, but at a different loss. This combinatorial nature of the problem makes it difficult to find the global optimum. We propose an efficient, globally optimal optimization technique. By exploiting the structure of the problem we are able to identify areas of the solution space that cannot contain the optimal solution and discard them early on. 
Our globally optimal algorithm is able to calibrate a few hundred E-SVMs quickly. In order to solve larger problems with thousands of E-SVMs, we present a simple modification of our exact algorithm to deliver high quality approximate solutions.

The rest of the paper is organized as follows. We start by reviewing related work in sec.~\ref{sec:rel_work}. Sec.~\ref{sec:form} introduces the formulation of our optimization problem, while sec.~\ref{sec:algo} presents our algorithm for efficiently finding the global optimal solution as well as its approximation.
We train EE-SVM on state-of-the-art CNN descriptors~\cite{girshick14cvpr} and present experiments on 10 classes of the ILSVRC 2014 dataset \cite{ilsvrc14} and on all 20 classes of PASCAL VOC 2007 \cite{everingham10ijcv} in sec.~\ref{sec:exp}. These experiments show that (i) our joint calibration procedure outperforms standard independent sigmoid calibration~\cite{MalisiewiczICCV11} on the task of classifying windows as belonging to an object class or not; and (ii) this translates to better object detection performance. Finally, we conclude in sec.~\ref{sec:concl}.
%\todo{code release?}

\vspace{-1mm}
\section{Related Work}
\label{sec:rel_work}
\vspace{-2mm}
In the machine learning literature, classifier calibration has been considered in the context of deriving probabilistic output for binary classifiers~\cite{platt1999, zadrozny01kdd, zadrozny01icml, niculescu2005icml} or multi-class classification~\cite{zadrozny02kdd,gronat13cvpr}. 
Multi-class problems are often cast as a series of binary problems (e.g. 1-vs-all) and ~\cite{zadrozny02kdd, MalisiewiczICCV11, aubry2014tog} showed that calibrating these binary classifier often leads to improved prediction.

The two most popular methods for calibrating binary classifiers are Platt scaling~\cite{platt1999} and isotonic regression~\cite{zadrozny01kdd}. They both fit a monotonic function of the classifier score to the empirical label probability, obtaining an estimate of the conditional probability of a class label given the score. Platt scaling~\cite{platt1999} uses a simple sigmoid function, while~\cite{zadrozny01kdd} employs a more flexible isotonic regression. In computer vision, Platt scaling is the most popular calibration tool~\cite{MalisiewiczICCV11,hoiem2008ijcv,endres13cvpr}.
We compare to both methods in sec.~\ref{sec:exp_gos}.

All these works~\cite{platt1999, zadrozny01kdd, zadrozny01icml, niculescu2005icml} assume that the set of positive training samples for each classifier is fixed and given beforehand, even if small.
In contrast, in the EE-SVM model, any positive sample can potentially be associated with any E-SVM. In the original EE-SVM paper~\cite{MalisiewiczICCV11} this was resolved in a greedy fashion, where each E-SVM was calibrated independently. The association of a positive sample to an E-SVM was resolved by comparing its uncalibrated E-SVM score to a fixed threshold.
Instead, we calibrate E-SVMs \emph{and} associate positive samples with them jointly over all positives and all E-SVMs. Our joint formulation (sec.~\ref{sec:form}) ensures that every positive is associated with at least one E-SVM, while the total number of false positives is minimized. As an additional benefit, this enables removing up to 25\% of redundant E-SVMs that are not associated with any positives after the global optimum is found.

%\todo{can we state a long list of papers using EE-SVM that uses the classic MalisiewiczICCV11 procedure?} % no, can't find many; most works use E-SVMs not in a max-ensemble :-(
Two interesting exceptions to the classic EE-SVM calibration procedure~\cite{MalisiewiczICCV11} were presented recently \cite{gronat13cvpr, aubry2014cvpr}. Gronat \etal \cite{gronat13cvpr} learns a per-location classifier for visual place recognition, while \cite{aubry2014cvpr} learns exemplar-based 3D ``chair'' representations.
Both works employ a calibration strategy based purely on negative samples, sidestepping the
association of positive samples to E-SVMs. % 'are unable to associate' sounds too strong: you could always associate after the fact, by picking the highest scoring E-SVM on a particular positive training sample
For completeness, we compare to \cite{aubry2014cvpr} in sec.~\ref{sec:exp_gos}. All techniques reviewed above calibrate each E-SVM independently.

\setlength{\belowcaptionskip}{0pt}
\begin{figure}
  \centering
    \includegraphics[width=\textwidth]{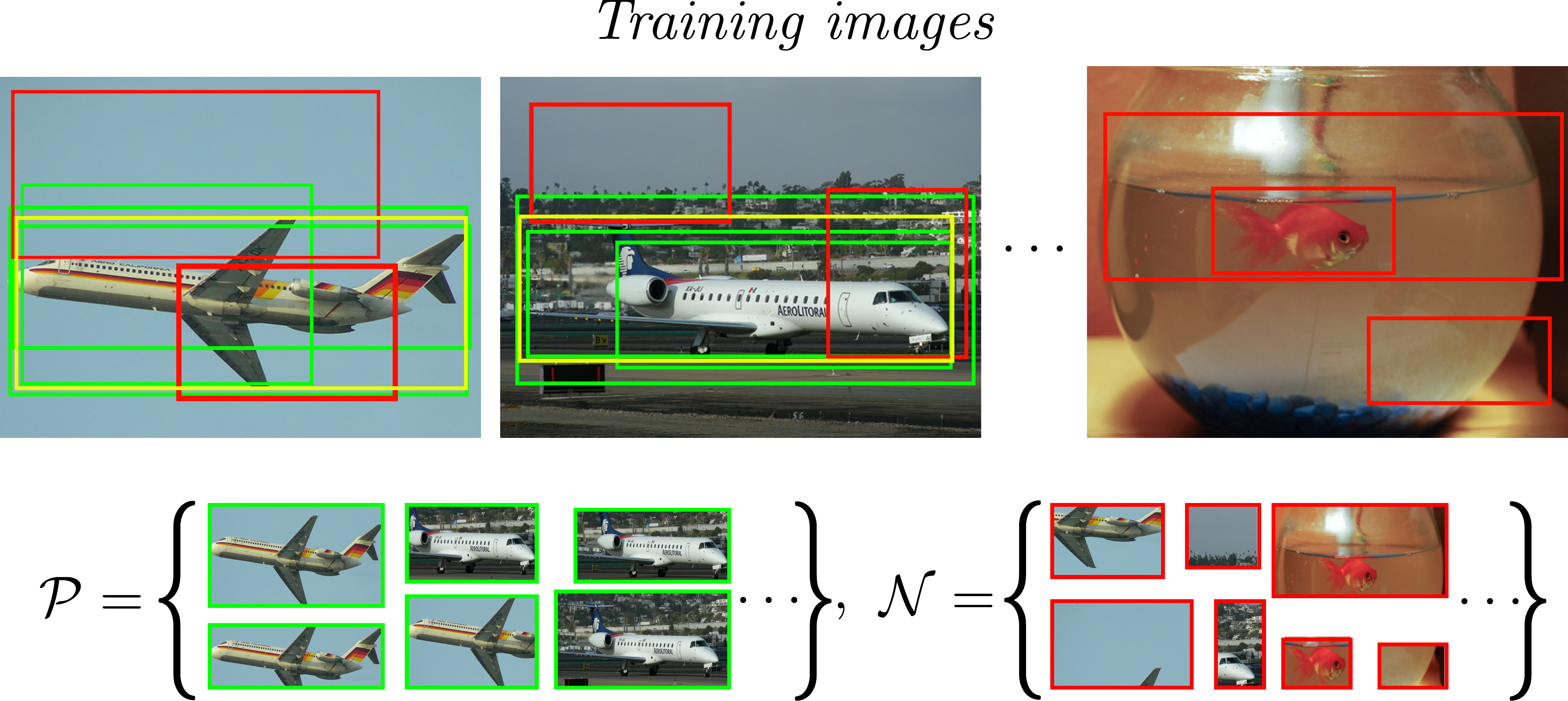}
      \caption{\small \it Example of window proposals used in our calibration technique. $\mathcal{P}$ is the set of positive windows ({\color[rgb]{0,1,0} $\Box$})  and $\mathcal{N}$ is the set of negative windows ({\color[rgb]{1,0,0} $\Box$}) in the training set. Finally, ({\color[rgb]{1,1,0} $\Box$}) indicates E-SVMs ground-truth bounding-box. A window is positive if it has an intersection-over-union $\geq 0.5$ with a ground-truth box~\cite{everingham10ijcv}.\vspace{-2mm}}
      \vspace{-2mm}
      \label{fig:windows}
\end{figure}

\begin{figure*}
  \centering
    \includegraphics[width=\textwidth]{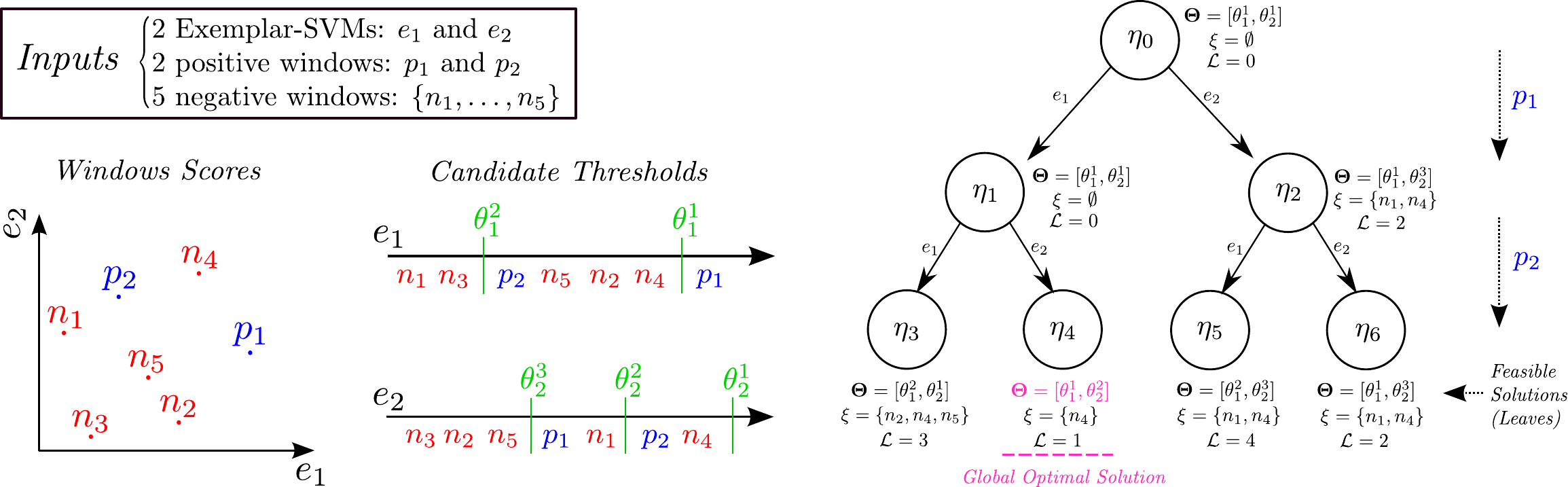}
      %\caption{\small \it Illustration of our joint calibration algorithm. Given the window scores, the candidate thresholds for the two E-SVMs $e_1$ and $e_2$ are $[\theta_1^1, \theta_1^2]$ and $[\theta_2^1, \theta_2^2, \theta_3^3]$, respectively. The tree represents the space of all possible solutions. The only feasible threshold configurations are those in the leaves. \vspace{-2mm}}
      \caption{\small \it Illustration of our joint calibration algorithm. \textbf{(left)} shows the window scores of the two E-SVMs $e_1$ and $e_2$. \textbf{(middle)} shows the candidate thresholds for these two E-SVMs. These are $[\theta_1^1, \theta_1^2]$ and $[\theta_2^1, \theta_2^2, \theta_3^3]$, respectively. Finally, \textbf{(right)} shows the tree representing the space of all possible solutions. Note how the only feasible threshold configurations are those in the leaves. \vspace{-4mm}}
      \vspace{-2mm}
      \label{fig:pipeline}
\end{figure*}

%\vspace{-1mm}
\section{Joint calibration formulation}
\label{sec:form}
\vspace{-2mm}

In many object detection pipelines~\cite{dalal05cvpr,girshick14cvpr,cinbis13iccv} a single linear classifier $w \in R^d$ is applied to all $K$ candidate windows ${\{x\}}_{i=1}^K$ in an image, where $x \in R^d$ is the window descriptor. 
The windows are then ranked according to the classifier score $w \cdot x$.
An EE-SVM, instead, contains $E$ classifiers: $\{w_j\}_{j=1}^E$.
The score of a window $x$ is defined as the highest score
among all classifiers applied to it:
\vspace{-1mm}
\begin{equation}
\vspace{-3mm}
S(w)  = \max_j(w_j \cdot x)
\end{equation}
%\vspace{-1mm}
Our goal is to find a threshold $\theta_j$ for each E-SVM $e_j$ such that
(i) all positive windows are scored positively by at least one E-SVM, and
(ii) the number of negative windows scored positively by any E-SVM is minimized.
A window $x$ is scored positively by E-SVM $e_j$ if $w_j \cdot x - \theta_j > 0$.
We formalize these criteria in an optimization problem:
\vspace{-2mm}
\begin{align}
\begin{split}
\min_{\mathbf{\Theta}=\{\theta_j\}_{j=1}^E} \overbrace{ \sum_{x \in \mathcal{N}}\mathds{1}[\max_j(w_j \cdot x -\theta_j)]}^{\mathcal{L}(\mathbf{\Theta})} \\
s.t. \; \mathds{1}[\max_{j}(w_j \cdot x-\theta_j)] > 0, \, \forall x \in \mathcal{P}
\end{split}
\label{eq:calib_formulation}
\end{align}
where $\mathds{1}$ is the indicator function and $\mathcal{P}$ and $\mathcal{N}$ are the sets of positive and negative windows in the training set (fig.~\ref{fig:windows}). We refer to the top term as the {\em loss function} $\mathcal{L}(\mathbf{\Theta})$ and the bottom terms as the {\em constraints}.

Calibration is performed by adjusting the thresholds $\mathbf{\Theta}$. 
Given a configuration of thresholds $\mathbf{\Theta} = [\theta_1, \theta_2, \dots \theta_E]$, the loss $\mathcal{L}(\mathbf{\Theta})$ counts the number of negative windows scored positively after the max operation. Each constraint $i$ ensures that a positive window $x_i$ is scored positively by at least one E-SVM. We refer to a configuration $\mathbf{\Theta}$ satisfying all the constraints as a \textit{feasible solution}. 
  
%\vspace{-1mm}
\section{Globally optimal and efficient solution}
\label{sec:algo}
\vspace{-2mm}
In this section we develop a computationally efficient algorithm to find the global optimal solution of (\ref{eq:calib_formulation}).
We start in sec.~\ref{sec:thresholds} by analysing the space of all possible solutions of (\ref{eq:calib_formulation}).
In sec.~\ref{sec:tree} we then introduce a data structure to represent this space, and finally in sec.~\ref{sec:efficient_search} we present an efficient algorithm to search this data structure for the globally optimal solution.

\subsection{Space of candidate thresholds}
\label{sec:thresholds}
\vspace{-2mm}

At first sight, (\ref{eq:calib_formulation}) appears to be a continuous optimization problem where each threshold can take any value in $\mathbb{R}$. %This  space is unfortunately infinitely large and practically unsolvable. 
However, since E-SVMs are evaluated only on a finite set of training windows, there exist an infinite set of \textit{equivalent} thresholds leading to the same loss.  For this reason, (\ref{eq:calib_formulation}) is in practice a discrete optimization problem. 

%Note that the infinite number of thresholds between two window scores are equivalent (fig.~\ref{fig:thresholdsInf}). In the rest of the work we are only going to consider thresholds in the middle of two scores.  

\begin{figure}[b]
\vspace{-3mm}
  \centering
    \includegraphics[width=\textwidth]{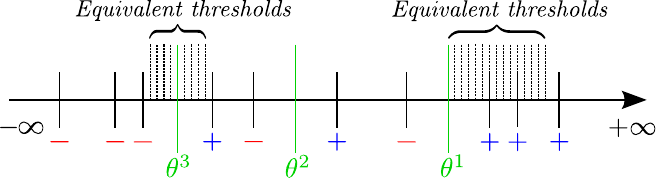}
      \caption{\small \it Candidate thresholds, given the scores on positive (\textcolor{blue}{+}) and negative (\textcolor{red}{-}) windows. The only thresholds worth considering according to (\ref{eq:calib_formulation}) are the ones between a negative and positive window. Of all equivalent thresholds between two window scores, we consider only the mean of the two scores.}
           \vspace{-2mm}
      \label{fig:thresholdsInf}
\end{figure}

Fig.~\ref{fig:thresholdsInf} shows an example. %The space of candidate thresholds of an E-SVM is defined by the number of windows evaluated.  \todo{too early to mention candidate thresholds; now move in by talking about which intervals of thresholds are equivalent, and so need to consider just one per interval}
Since each constraint in (\ref{eq:calib_formulation}) evaluates one positive windows, an E-SVM needs at most $P+1$ thresholds to satisfy each of them (fig.~\ref{fig:pipeline}), where $P=|\mathcal{P}|$.
Furthermore, considering a threshold between two positive samples is not necessary, because the loss only changes when new negative samples are scored positively. 
%\todo{careful 'positive score' does not mean 'score of a positive sample'. I fixed it in the prev sentence; please fix everywhere}
%
The only thresholds worth considering are those between the score of a negative sample and a positive (not the reverse, fig.~\ref{fig:thresholdsInf}).
We denote the set of candidate thresholds for an E-SVM $e_j$ as $[\theta_j^1, \theta_j^2, \dots \theta_j^{M_j}]$, where $1 \leq M_j \leq P+1$ and  $\theta_j^a < \theta_j^b$,  for $\forall a,b : 1 \leq a < b \leq M_j$. 
By construction, the lowest threshold $\theta_j^{M_j}$ satisfy all the constraints in (\ref{eq:calib_formulation}). 
%
%Fig.~\ref{fig:pipeline} shows an example. 

To conclude, the number of candidate thresholds for an individual E-SVM is relatively small (at most $P+1$), but the joint space of E-SVMs thresholds is nonetheless huge. In the worse case scenario (all E-SVMs have $P$ candidate thresholds), there are $P^E$ threshold configurations, many of which are not a feasible solution to (\ref{eq:calib_formulation}). In the next section we present a data structure to enumerate all these configurations and highlight the feasible solutions.

\subsection{Exhaustive search tree}
\label{sec:tree}
\vspace{-2mm}

We represent the space of all possible solutions as a search tree (fig.~\ref{fig:pipeline}).
\vspace{-1mm}
\begin{mydef} {\bf (Search Tree)}
Our search tree $T$ is a perfect k-ary tree: a rooted tree where every internal node has exactly k children and all leaves are at the same depth $h$. Each node $\eta$ contains a configuration $\mathbf{\Theta} = [\theta_1, \theta_2, \dots \theta_E]$ of thresholds. 
\end{mydef}
\vspace{-1mm}
A configuration $\mathbf{\Theta}$ at node $\eta$ is used to compute the loss $\mathcal{L}(\mathbf{\Theta})$ by counting how many negative windows  are scored positively according to (\ref{eq:calib_formulation}). The root node has the configuration $\mathbf{\Theta} = [\theta_1^1, \theta_2^1, \dots \theta_E^1]$ of all tightest threshold for each E-SVM.
We denote the set of false positives at a node $\eta$ as $\xi_\eta$ and $\mathcal{L}(\mathbf{\Theta_\eta}) = |\xi_\eta|$. Note how the root has $\xi = \emptyset$.
%In our representation we have $h=P$ and $k=E$. As shown in fig.~\ref{fig:pipeline}, each level $i$ of the tree, for $1 \leq i \leq P$, corresponds to a positive window $x_i$ and each edge $j$, for $1 \leq j \leq E$, corresponds to an E-SVM $w_j$.

In our representation we have $h=P$ and $k=E$. %Each level of the tree corresponds to a positive window and each edge corresponds to an E-SVM.
Each level $l$ of the tree corresponds to a positive window $p_l$, and each edge corresponds to an E-SVM $e_j$. An edge $e_j$ indicates that $e_j$ is responsible for $p_l$ and it should score it positively, hence satisfying one constraint of (\ref{eq:calib_formulation}). %At each level $l$, the path down the tree chooses which E-SVM should be responsible for $p_l$.
Given an E-SVM $e_j$ and its current threshold $\theta_j$, the edge lowers the threshold so that $w_j \cdot x_l-\theta_j > 0$ (if this condition is already satisfied then the threshold does not change). Lowering the threshold {\em might} increase the loss, but not necessarily.
Lowering the threshold will make E-SVM $e_j$ score positively some negative windows, but these affect the loss only if they were not already scored positively by another E-SVM.

The deeper the level, the more constraints of (\ref{eq:calib_formulation}) are satisfied. By construction,
the configuration of thresholds at a leaf satisfy all constraints, and the set of all leaves represent the set of all feasible solutions. Also note that the number of false positives always increases (or remains the same) along a path from the root to a leaf: given a node $\eta$ and any child $\eta'$, we have that $\mathcal{L}(\mathbf{\Theta}_{\eta}) \leq \mathcal{L}(\mathbf{\Theta}_{\eta'})$.

\subsection{Efficient search}
\label{sec:efficient_search}
\vspace{-2mm}

In this section we find the global optimal solution to (\ref{eq:calib_formulation}) by searching the tree.
Exhaustive search is computationally prohibitive even for small problems with few E-SVMs and positive windows, as the total number of nodes in our tree is $(E^{P+1}-1)/(E-1)$ (with $E$ the number of E-SVMs and $P$ the number of positive windows).

%The algorithm unfortunately doesn't allow for pruning by optimality: 
%When the search reaches a leaf of the tree, we do not have any optimality condition to check if its feasible solution is optimal or not. In order to guarantee the optimality of a given feasible solution, we need ``to compare'' it  against all other feasible solutions. 

The key to our efficient algorithm is to prune the tree by iteratively removing subtrees which cannot contain the global optimal solution.
In the following paragraphs we present several observations that enable to drastically reduce the space of solutions to consider. The last paragraph presents the actual search algorithm.

%  ------------------- OBSERVATION 1 

\begin{observation}
{\bf (Pruning by bound)}. If $\eta$ is a leaf and $\eta'$ is a node not on the path from the root to $\eta$, and $\mathcal{L}(\mathbf{\Theta}_{\eta}) \leq \mathcal{L}(\mathbf{\Theta}_{\eta'})$, then the subtree rooted at $\eta'$ cannot contain a better solution than $\eta$ and can be discarded.
\end{observation}

The key intuition is that the loss can only increase with depth.
The observation leads to two cases.
First, if $\mathcal{L}(\mathbf{\Theta}_{\eta}) < \mathcal{L}(\mathbf{\Theta}_{\eta'})$, then $\eta'$ cannot lead to an optimal solution since its loss is already higher than an already found feasible solution.
Second, if $\mathcal{L}(\mathbf{\Theta}_{\eta}) = \mathcal{L}(\mathbf{\Theta}_{\eta'})$, $\eta'$ could lead to an optimal solution, but it would be equivalent to the one in $\eta$.
In both cases we can discard the subtree rooted at $\eta'$, as we are interested in finding only one optimal solution.

\begin{figure}
  \centering
    \includegraphics[width=.9\textwidth]{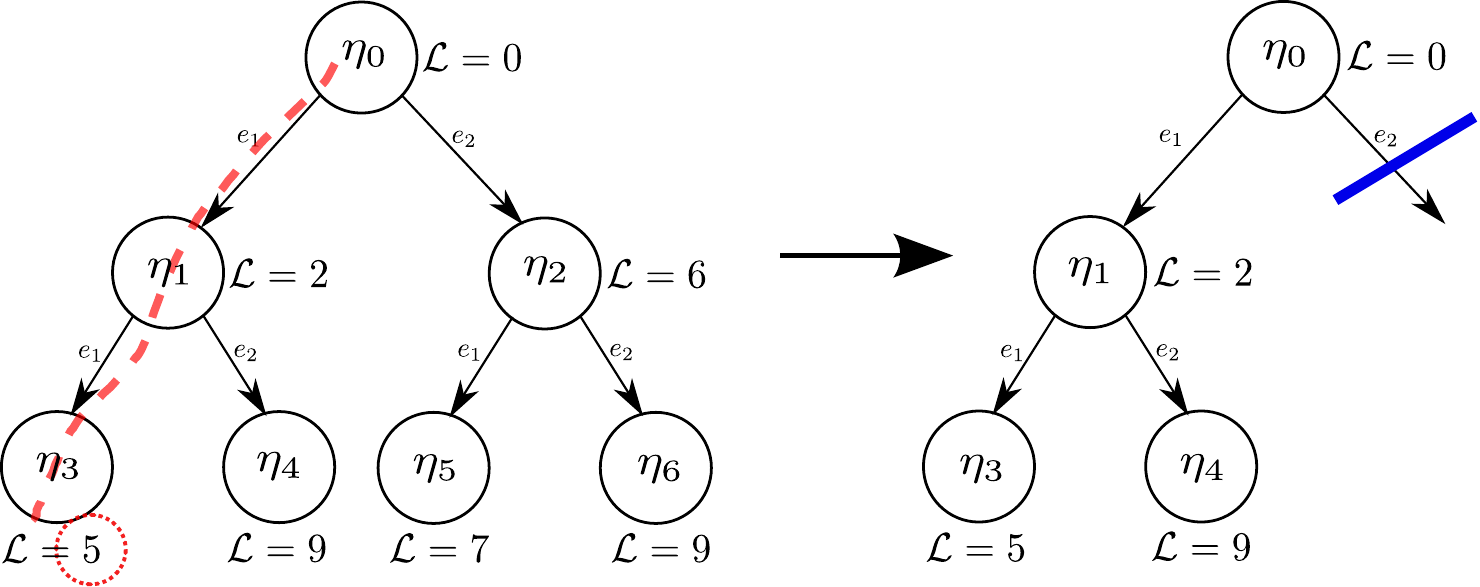}
    %\includesvg[pdf, width=\textwidth]{figures/pruning_by_bound}
      \caption{\small \it Example of pruning by bound. Since $\mathcal{L}(\mathbf{\Theta}_{\eta_3}) =5 < \mathcal{L}(\mathbf{\Theta}_{\eta_2}) = 6$, the subtree rooted at $\eta_2$ cannot contain an optimal solution.
  %: any solution in it has a loss $\geq 6$, which is higher than the feasible solution in  $\eta_3$. Hence, there is no need to search it.} %Once the feasible solution in $\eta_3$ is discovered, the unexplored branch rooted at $\eta_2$ can be cut.
    }
      \vspace{-1mm}
      \label{fig:pruning_bound}
\end{figure}

Consider the example in fig.~\ref{fig:pruning_bound}. Since $\mathcal{L}(\mathbf{\Theta}_{\eta_3}) =5 < \mathcal{L}(\mathbf{\Theta}_{\eta_2}) = 6$, the subtree rooted at $\eta_2$ cannot contain an optimal solution: any solution in it has a loss $\geq 6$, which is higher than the feasible solution in  $\eta_3$.

%  ------------------- OBSERVATION 2

\begin{observation} 
{\bf (Pruning by equivalence)}. If two nodes $\eta$ and $\eta'$ have the same parent and $\xi_\eta = \xi_{\eta'}$, then they are equivalent and only one subtree needs be searched.
\end{observation}

\begin{figure}
  \centering
    \includegraphics[width=.9\textwidth]{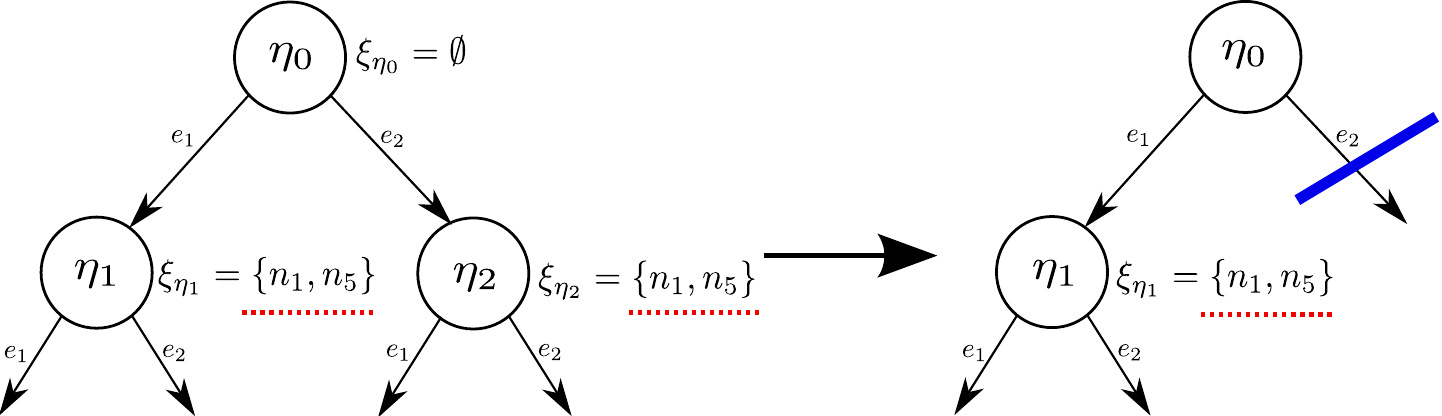}
    %\includesvg[pdf, width=\textwidth]{figures/pruning_by_equivalence}
      \caption{\small \it Example of pruning by equivalence. Since the two nodes $\eta_1$ and $\eta_2$ have the same parent and $\xi_{\eta_1} = \xi_{\eta_2}$, they are equivalent and only one subtree needs be searched.}
      \vspace{-1mm}
      \label{fig:pruning_equivalence}
\end{figure}

The key intuition is that the loss in (\ref{eq:calib_formulation}) only increases when \textit{new} negative windows are scored positively. 
Consider the example in fig.~\ref{fig:pruning_equivalence}, where $\xi_{\eta_1} = \xi_{\eta_2}$ and $\eta_1$ and $\eta_2$ have the same parent (i.e., they satisfy the same constraints of (\ref{eq:calib_formulation})). $\eta_1$ has $\mathbf{\Theta} = [\theta_1^2, \theta_2^1]$ because the edge from $\eta_0$ to $\eta_1$ adjusted the threshold of $e_1$. Note, however, that this configuration can be changed into $[\theta_1^2, \theta_2^2]$ without increasing the loss, as they are equivalent solutions. Because of this equivalence, both subtrees lead to equivalent feasible solutions and only one of them needs be searched.

%  ------------------- OBSERVATION 3

\begin{observation} 
{\bf (Reducing tree depth)}. Given the root configuration $\mathbf{\Theta} = [\theta_1^1, \theta_2^1, \dots \theta_E^1]$, there might exist some $x \in \mathcal{P} : \max_j (w_j \cdot x-\theta_j^1) > 0$. Since $\mathbf{\Theta}$ already satisfies the constraint for these positives at zero cost, these can be eliminated right away, reducing the depth of the tree.
\end{observation}

Consider the example in fig.~\ref{fig:pipeline}. Initially, $\mathbf{\Theta} = [\theta_1^1, \theta_2^1]$. This configuration already scores $p_1$ positively. Whatever optimal solution $\mathbf{\Theta}$ the tree retrieves, $p_1$ will always be scored positively by at least E-SVM $e_1$. Hence, we can  eliminate it from the tree to reduce its depth.

\begin{figure}
  \centering
    \includegraphics[width=1\textwidth]{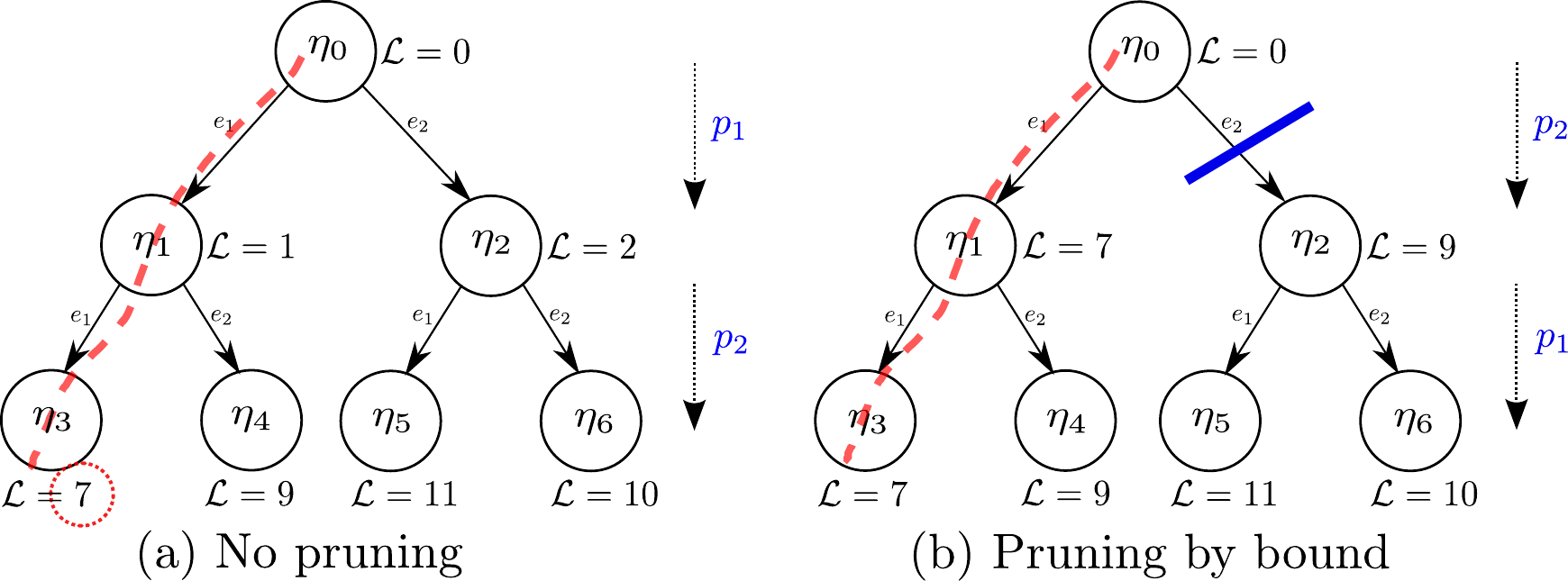}
    %\includesvf[pdf, width=1.1\textwidth]{figures/fast_pruning_by_bound}
      \caption{\small \it Efficient pruning by bound can be achieved by sorting the positive windows by decreasing difficulty. \vspace{-2mm}}
      \vspace{-1mm}
      \label{fig:fast_pruning_bound}
\end{figure}

%  ------------------- OBSERVATION 4 

\begin{observation} 
{\bf (Order of positive windows)}.
By sorting the positive windows by decreasing difficulty, pruning by bound can discard larger subtrees. The difficulty of a positive window $x$ is measured as $\min_j \delta(e_j, x)$.
%$\mathbf{\Theta}_j^* = [\theta_1^1, \dots \theta_j^*, \dots \theta_E^1]$ and $\theta_j^*$ is the highest threshold of E-SVM $e_j$ scoring $x_i$ positively. \todo{can we rewrite this so it only depends on one threshold? Why do we need to invoke all E-SVMs?}
\end{observation}

where $\delta(e_j, x)$ counts how many false positives $e_j$ produces by scoring the positive sample $x \in \mathcal{P}$ positively.

The key intuition is that it is better to prune subtrees rooted at the higher levels of the tree, as they contain more nodes.
This can be achieved by placing difficult positive windows up in the higher levels. Some positive windows lead to constraints intrinsically more difficult to satisfy than others, as any E-SVM asked to satisfy it would score positively many negatives as well, and hence incur a large loss.
By sorting the tree levels according to the difficulty of positive windows,  it is likely that the loss of many high-level configurations is higher than a previously found feasible solution, and therefore can be pruned (observation 1).

Consider the example in fig.~\ref{fig:fast_pruning_bound}. Tree (a) evaluates first $p_1$ and then $p_2$, while (b) does the opposite. Tree (a) cannot be pruned by observation 1, but tree (b) can.

In sec.~\ref{sec:pruning} we evaluate experimentally how effective the above pruning techniques are on various real EE-SVM calibration problems. 

\vspace{-4mm}
\paragraph{Search algorithm.}
\label{list_functions}
We present here a depth-first search algorithm to efficiently find the global optimal solution, based on the above observations (Algo.~\ref{algo}).

The algorithm works as follows.
The initial configuration of thresholds $\mathbf{\Theta}$ is the one from the root node \textit{\small (line 1)}.
As preprocessing, the algorithm starts by reducing the tree depth using observation 3 \textit{\small (line 2)} and re-ordering the positive windows using observation 4 \textit{\small (line 3)}.
In the first step, it does a depth-first search until it reaches a leaf $\eta$ and finds a first feasible solution $\mathbf{\Theta}$ \textit{\small (line 4)}. During this traversal, when going down a level, the algorithm always chooses the edge leading to the smallest loss.
Next, the algorithm continues by going up \textit{\small (line 6)} and down \textit{\small (line 10)} the tree.
When visiting a node, the algorithm tries to prune as many children subtrees as it can using observation 1 \textit{\small (line 8)} and observation 2 \textit{\small (line 9)}.
When the algorithm reaches a leaf then this must contain a better solution than the current one $\mathbf{\Theta}$ (in term of the loss (\ref{eq:calib_formulation})). Hence, it updates $\mathbf{\Theta}$ \textit{\small (line 13)}. The algorithm continues until all nodes have been visited or pruned.
The final $\mathbf{\Theta}$ is the global optimum of (\ref{eq:calib_formulation}).

\vspace{-0.5mm}
\subsection{Approximate search}
\label{sec:approx}
\vspace{-2mm}
Above we presented an efficient algorithm that guarantees global optimality. If we relax the global optimality requirement, we can improve efficiency even further.
Our method follows a depth-first search and as soon as it reaches a leaf, then it finds a feasible solution. This happens periodically during the execution of the algorithm, as better and better leaves are found while the tree is searched. 
This behaviour makes our method an \textit{any-time} algorithm~\cite{zilberstein96ai, gogate2004auai}. After a short period required to reach the first leaf, we can terminate it at any time and it will return the best feasible solution it has found so far (although not necessarily the globally optimal one).
This simple observation enables to employ our method, essentially unchanged, to find approximate solutions as well.

\begin{algorithm} [t]
\renewcommand{\algorithmicrequire}{\textbf{Input:}}
\renewcommand{\algorithmicensure}{\textbf{Output:}}
\newcommand{\BlankLine}{\vskip 1ex}
\caption{Our Efficient Search Algorithm} 
\label{algo}
\begin{algorithmic}[1]
\BlankLine
\REQUIRE search tree $T$

\ENSURE global optimal $\mathbf{\Theta}$
\BlankLine

%\STATE $\beta \leftarrow$ \textsc{Compute$\,$Initial$\,$Upper$\,$Bound}
\STATE  $\mathbf{\Theta} \leftarrow [\theta_1^1, \theta_2^1, \dots \theta_E^1]$
\STATE $T  \leftarrow$ \textsc{Reduce$\,$Tree$\,$Depth($T$, $\mathbf{\Theta}$)}
\STATE $T  \leftarrow$ \textsc{Reorder$\,$Positive$\,$Windows($T$, $\mathbf{\Theta}$)}
\STATE $\eta$, $\mathbf{\Theta}$, $\mathcal{L} \leftarrow$ \textsc{Depth$\,$First$\,$Search($T$, $\mathbf{\Theta}$)}
\WHILE{$T$ not fully searched}
\STATE $\eta \leftarrow$ \textsc{Go$\,$Up$\,$One$\,$Level} $(T, \eta)$
\WHILE{$\neg$ \textsc{isLeaf($\eta$)} $\land$ $\neg$ \textsc{hasChild($\eta$)}}
   \STATE $T \leftarrow$ \textsc{Prune$\,$By$\,$Bound} $(T, \eta, \mathcal{L})$
   \STATE $T \leftarrow$ \textsc{Prune$\,$By$\,$Equivalence} $(T, \eta, \mathbf{\Theta})$
   \STATE $\eta \leftarrow$ \textsc{Go$\,$Down$\,$One$\,$Level} $(T, \eta)$
\ENDWHILE
\IF{\textsc{isLeaf($\eta$)}}
\STATE $\mathbf{\Theta}$, $\mathcal{L} \leftarrow$ \textsc{Get$\,$Solution($\eta$)}
\ENDIF
\ENDWHILE

\BlankLine
\STATE return $\mathbf{\Theta}$
\BlankLine
\end{algorithmic}
\end{algorithm}

\section{Experiments} \label{sec:exp}
\vspace{-1mm}

\begin{table*}[t!]
\small \centering
\resizebox{\columnwidth}{!}
{
\begin{tabular}{ | l | c c c c c c c c c c | c | } % {| >{\centering\arraybackslash}m{1in} | >{\centering\arraybackslash}m{1in} |}
\hline
{\bf ILSVRC 2014 - trained on \texttt{Val$\mathbf{_1}$}}& Airplane & Bagel & Baseball & Bear & Butterfly & Koala & Ladle & Printer & Sheep & Violin & mean\\\hline
\it  Recall & \it 94.1  & \it 90.1  & \it 97.9  & \it 93.1  & \it 97.5  & \it 96.6  & \it 79.5  & \it 88.5  & \it 98.9  & \it 85.3  & \it 92.2 \\\hdashline
%& \it (899/955) & \it (1795/1992) & \it (760/776) & \it (2187/2349) & \it (3584/3675) & \it (623/645) & \it (660/830) & \it (669/756) & \it (1311/1326) & \it (424/497) \\\hdashline
EE-SVM no calibration & 180124 & 171966 & 499056 & 33664 & 163727 & 80145 & 250602 & 221117 & 308458 & 37519 & 195k\\
EE-SVM indep. sigmoid calibration \cite{MalisiewiczICCV11} & 119552 & 42165 & 234734 & 77099 & 53986 & 13017 & 56616 & 88390 & 86507 & 33266 & 80k\\
EE-SVM joint calibration & 65182 & 28460 & 180658 & 22694 & 53927 & 10746 & 87513 & 36573 & 55923 & 32570 & 57k \\
EE-SVM joint calibration w/ sigmoid & 64996 & 28140 & 168129 & 22876 & 53867 & 10722 & 87329 & 35803 & 50064 & 33899 & \bf 55k \\\hdashline
EE-SVM indep. isotonic regression \cite{zadrozny02kdd} & 54173 & 23625 & 268302 & 16424 & 43580 & 13507 & 60285 & 55129 & 76997 & 13814 & 63k \\
EE-SVM indep. affine calibration \cite{aubry2014cvpr} & 51905 &  24507 & 224003 & 18978 & 37483 & 9947 & 67288 & 86757 & 110909 & 18218 & 65k \\\hdashline
Single Linear-SVM (R-CNN) \cite{girshick14cvpr} & 102676 & 335122 & 711185 & 109480 & 63849 & 305931 & 322332 & 469777 & 979131 & 121050 & 341k \\\hline
\end{tabular}}
\caption{\small \it \textbf{Window classification - False positives at test recall.} Results on a subset of ILSVRC 2014 \texttt{Val$_2$} (all positive windows and one million randomly sampled negative ones). We use the optimal thresholds found by our algorithm (sec.~\ref{sec:algo}) to compute recall on \texttt{Val$_2$}. This is the percentage of positive windows scored positively by our jointly calibrated EE-SVM. The table entries show the number of false positives produced in order to reach that recall level. Each row corresponds to a different method (ours are marked `joint calibration'). \vspace{-1mm}}
%\caption{\small \it \textbf{Window classification - False positives at test recall.} Results on ILSVRC 2014 \texttt{Val$_2$}. We use the optimal thresholds found by our algorithm (sec.~\ref{sec:algo}) to compute recall as the percentage of positive windows scored positively by the ensemble on \texttt{Val$_2$}. The entries in the table represent the number of false positives produced by the different methods.}
\vspace{-2mm}
\label{table:exp1}
\end{table*}

\begin{table*}[t!]
\footnotesize \centering
\resizebox{\columnwidth}{!}
{
\begin{tabular}{ | l | c c c c c c c c c c | c | } % {| >{\centering\arraybackslash}m{1in} | >{\centering\arraybackslash}m{1in} |}
\hline
{\bf ILSVRC 2014 - trained on \texttt{Val$\mathbf{_1}$}}& Airplane & Bagel & Baseball & Bear & Butterfly & Koala & Ladle & Printer & Sheep & Violin & mAP\\\hline
EE-SVM indep. sigmoid calibration \cite{MalisiewiczICCV11} &  42.8 & 39.7 & 63.3 & 58.7 & 60.8 & 58.2 & 4.5 & 29.0 & 49.5 & 20.3 & 42.7\\
EE-SVM joint calibration w/ sigmoid &  43.3 &  40.1 &  66.5 &  60.6 &  63.9 &  61.1 &  5.0 &  31.6 &  55.1 &  22.9 & \bf 45.0\\\hdashline
EE-SVM indep. isotonic regression \cite{zadrozny02kdd} & 44.6 & 42.4 & 61.4 & 59.3 & 63.8 & 63.2 & 5.7 & 22.5 & 50.9 & 23.2 & 43.7 \\
EE-SVM Indep. affine calibration \cite{aubry2014cvpr} & 45.6 & 42.0 & 64.1 & 59.2 & 62.6 & 62.2 & 6.3 & 22.9 & 50.4 & 25.9 & 44.1 \\\hdashline
Single Linear-SVM (R-CNN) \cite{girshick14cvpr} & 47.9 & 36.9 & 65.0 & 60.9 & 66.7 & 63.4 & 5.4 & 24.4 & 50.6 & 19.1 & 44.0 \\\hline
\end{tabular}}
\caption{\small \it \textbf{Window classification - Average precision.} Results on a subset of ILSVRC 2014 \texttt{Val$_2$} (same data as table~\ref{table:exp1}). \vspace{-3mm}}
\vspace{-2mm}
\label{table:exp2}
\end{table*}

\subsection{Datasets} \label{sec:dataset}
\vspace{-2mm}
We present experiments on ILSVRC2014~\cite{ilsvrc14} (sec.~\ref{sec:exp_gos},~\ref{sec:exp_approx}) and PASCAL VOC 2007~\cite{everingham10ijcv} (sec.~\ref{sec:exp_approx}).
ILSVRC2014~\cite{ilsvrc14} contains 200 classes annotated by bounding-boxes. In our experiments we randomly sampled 10 classes: \textit{airplane, bagel, baseball, bear, butterfly, koala, ladle, printer, sheep} and \textit{violin}. 
Following~\cite{girshick14cvpr} we consider three disjoint subsets of the data: \texttt{train}, \texttt{val$_1$} and \texttt{val$_2$}. Since annotations for the test set are not released, we measure performance on \texttt{val$_2$}. We use \texttt{val$_1$} and \texttt{train} for training. In total, these sets contain $>$80k images.
%We run our globally optimal algorithm on the \texttt{val$_1$} set~\cite{girshick14cvpr}, and our approximate algorithm on the much larger set union of \texttt{val$_1$}, \texttt{train$_{13}$} and \texttt{train$_{14}$}, where \texttt{train$_{13}$} is the part of the training set shared with the ILSVRC2013 detection dataset \cite{ilsvrc13} while \texttt{train$_{14}$} is unique to ILSVRC2014.

PASCAL VOC 2007~\cite{everingham10ijcv} contains 20 classes annotated by bounding-boxes. In our experiments we evaluate on all 20 classes. We use the subset \texttt{trainval} for training and we measure performance on \texttt{test}. In total, these sets contain about 10k images.

\subsection{Settings}
\vspace{-1mm}
\paragraph{Object proposals and features.}
We generate class-independent object proposals using~\cite{uijlings13ijcv}. Given an image, this produces a small set of a few thousand windows likely to cover all objects. 
We then extract state-of-the-art CNN descriptors of 4096 dimensions for these proposals, as in~\cite{girshick14cvpr}. These descriptors are the output of a convolutional neural network (CNN) initially trained for image classification~\cite{krizhevsky12nips, jia13caffe} and then fine-tuned for object detection~\cite{girshick14cvpr} (on \texttt{val$_1$} of ILSVRC2014, or on \texttt{trainval} of PASCAL VOC 2007).

\vspace{-3mm}
\paragraph{EE-SVM.}
We learn a separate window classifier $e$ for each instance of an object in the training set. We set $C=10^{-4}$ and we mine hard negatives from 2000 random training images. In our experiments we observed that mining more images did not bring a significant improvement.

\vspace{-3mm}
\paragraph{Calibration data.}
For each class we define $\mathcal{P}$ as the set of all positive training windows. A window is considered positive if it has intersection-over-union (IoU) \cite{everingham10ijcv} $\geq 0.5$ with a ground-truth bounding-box of that class. Moreover, $\mathcal{N}$ contains negative windows that overlap $\le 0.2$.
All calibration methods below are trained from this data.

\vspace{-3mm}
\paragraph{Independent sigmoid calibration.}
As a baseline we compare against the standard technique of Malisiewicz \etal~\cite{MalisiewiczICCV11}.
%This is the most used EE-SVM calibration procedure \todo{cite them all} and our method aims at improving it. % we could not find many papers doing it actually ;)
%
It operates in two steps. In step (1) it runs each E-SVM detector separately on a validation set, applies non-maximum suppression,
and then eliminates all detections scoring below the $-1$ margin. All remaining detections are considered positives if they belong to $\mathcal{P}$, or negatives if they belong to $\mathcal{N}$. Note how this arbitrarily defines which positive training samples to associate with a certain E-SVM.
In step (2), it then fits a logistic sigmoid to these data samples.

%this expensive procedure is necessary to form the subsets of positive and negative windows associated to an E-SVM.  \todo{not easier, but better?}.

\vspace{-3mm}
\paragraph{Our joint calibration.}
Our joint calibration also operates in two steps. 
In step (1), instead of arbitrarily defining the positive training samples, our technique use the thresholds found by our algorithm (sec.~\ref{sec:algo}) to associate positive samples to E-SVMs, which is the core underlying problem at the heart of such calibration. 
More specifically, for an E-SVM $e_j$, we consider as positives all windows $x \in \mathcal{P} : w_j \cdot x > \theta_j$, and as negatives all windows in $\mathcal{N}$.
Step (2) of our procedure is then the same as in~\cite{MalisiewiczICCV11}, but thanks to these optimal assignments, we fit better sigmoids.

We experimentally evaluate performance after each step. We refer to the output of step (1) as \textit{joint calibration}, and to the output of step (2) as \textit{joint calibration with sigmoid}.
As step (1) fits thresholds, it results in binary classification of test windows, while step (2) produces a continuous score which can be used for later processing stages (e.g. non-maximum suppression for object detection).

%Compared to the individual sigmoid calibration procedure, our algorithm fits thresholds, which results in binary classification of test windows. For clarity, in our experiments we refer to this first step as \textit{joint calibration}. 
%
%A sigmoid instead is a non-linear transformation of the scores which acts as a soft-threshold, stretching their range into $[0,1]$. Such continuous scoring function is sometimes useful for later processing stages (e.g. non-maxima suppression for detection). We extend our method by simply fitting a logistic sigmoid to the samples described above. In our experiments we refer to this as \textit{joint calibration with sigmoid.}

\vspace{-3mm}
\paragraph{Other independent calibration techniques.}
For completeness, we also compare against two independent calibration techniques not commonly used for EE-SVM: isotonic regression \cite{zadrozny02kdd} and the recent affine calibration \cite{aubry2014cvpr}. 
{\em Isotonic regression} fits a piecewise-constant non-decreasing function to the output of each E-SVM. We used the code of \cite{IsoRegCode} to train the function parameters on $\mathcal{P}$ and $\mathcal{N}$.
{\em Affine calibration} fits an affine transformation to the output of each E-SVM. As in \cite{aubry2014cvpr}, we train the affine parameters on 200k randomly sampled negative windows from $\mathcal{N}$.

\vspace{-3mm}
\paragraph{Single Linear-SVM (R-CNN).}
Finally, we provide results for the state-of-the-art R-CNN object detection model~\cite{girshick14cvpr}.
The sole purpose is to compare performance to EE-SVM on CNN features, as previous EE-SVM works typically use weaker HOG features~\cite{MalisiewiczICCV11, Shrivastava2011sgrapha, singh12eccv, aytar12bmvc, endres13cvpr, dong13cvpr, tighe13cvpr, juneja13cvpr, vezhnevets14cvpr}. However, as it consists of a single linear SVM per class, R-CNN cannot associate training exemples to objects detected in test images. Hence, it is not suitable for annotation transfer. We trained the model using the code and parameters of \cite{girshick14cvpr}. Note how this uses the same object proposals and features as our EE-SVM models.

% Compared to the individual calibration procedure, our learning algorithm only fits thresholds. We use these thresholds to define the E-SVMs' calibration positives and we then use these positives to fit a logistic function in a similar way of the individual calibration.
%More precisely, after obtaining a set of non-redundant detections using non-maxima suppression, we treat as positives for E-SVM$_j$ only those windows with  $IoU \ge 0.5$ and score higher than the threshold $\theta_j$. This is a major difference compared to the individual calibration. We use as negatives all windows with $IoU \le 0.2$ and we fit a logistic function to these scores.

\begin{figure*}
  \centering
    \includegraphics[width=\textwidth]{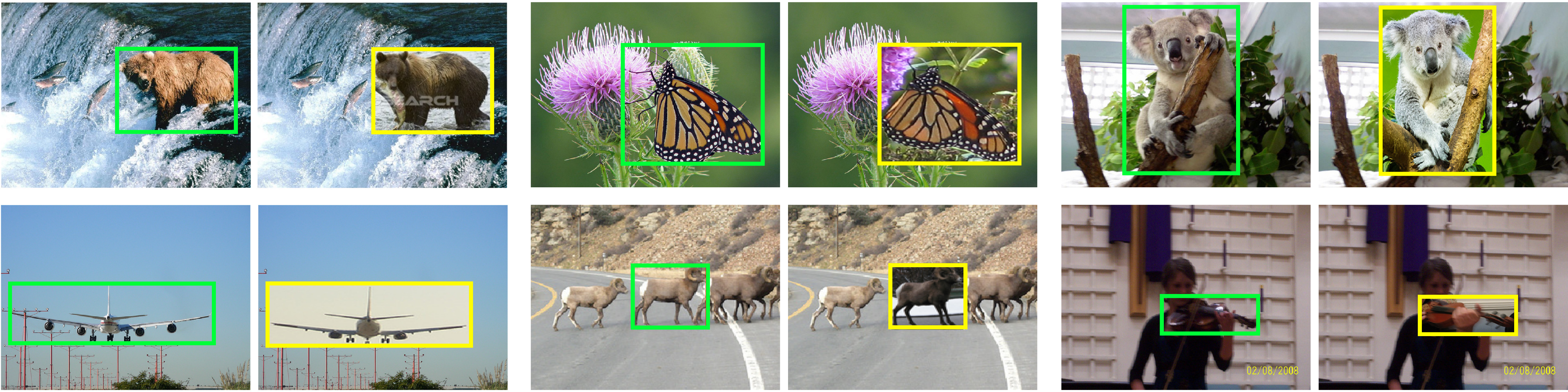}
      \caption{\small \it \textbf{Association between detected objects and training exemplars.} Our globally optimal joint calibration is good at transferring annotations from exemplars onto test windows. In these figure we show detections (green) and their associated training exemplar superimposed on them (yellow). \vspace{-2mm}}
      \label{fig:ann_transfer}
      \vspace{-2mm}
\end{figure*}

\begin{table*}[t!]
\footnotesize \centering
\resizebox{\columnwidth}{!}
{
\begin{tabular}{ | l | c c c c c c c c c c | c | } % {| >{\centering\arraybackslash}m{1in} | >{\centering\arraybackslash}m{1in} |}
\hline
{\bf ILSVRC 2014 - trained on \texttt{Val$\mathbf{_1}$ }}& Airplane & Bagel & Baseball & Bear & Butterfly & Koala & Ladle & Printer & Sheep & Violin & mAP\\\hline
EE-SVM indep. sigmoid calibration \cite{MalisiewiczICCV11} &  43.3 &  11.9 & 27.2 &  45.2 & 51.1 &  46.3 & 0.6 & 8.4 & 31.4 & 7.4 & 27.3 \\
EE-SVM joint calibration w/ sigmoid&  46.2 & 10.1 &  41.3 &  44.7 &  66.8 & 41.4 &  1.0 &  10.8 &  34.3 &  9.5 & 30.6\\\hdashline
EE-SVM indep. isotonic regression \cite{zadrozny02kdd} & 34.9 & 13.0 & 31.4 & 36.1 & 59.2 & 48.6 & 0.6 & 13.0 & 31.0 & 3.8 & 27.2 \\
EE-SVM indep. affine calibration \cite{aubry2014cvpr} & 45.8 & 10.3 & 41.0 & 44.1 & 66.1 & 39.5 & 0.8 & 11.4 & 35.9 & 9.6 & 30.5 \\\hdashline
Single Linear-SVM (R-CNN) \cite{girshick14cvpr} & 49.0 & 17.4 & 45.4 & 53.3 & 69.6 & 61.4 & 2.8 & 18.9 & 41.5 & 11.0 & 37.0 \\\hline
\end{tabular}}
\caption{\small \it \textbf{Object detection - Average precision.} Results on ILSVRC 2014 \texttt{Val$_2$}. \vspace{-3mm}}
\vspace{-2mm}
\label{table:exp3}
\end{table*}

\subsection{Globally optimal joint calibration} \label{sec:exp_gos}
\vspace{-2mm}

We evaluate here our globally optimal joint calibration technique on ILSVRC2014~\cite{ilsvrc14}. We train E-SVMs on \texttt{val$_1$} for the 10 classes listed in sec.~\ref{sec:dataset}. Each class has between 30 and 140 E-SVMs and between 500 and 3600 positive windows $\mathcal{P}$. 
We evaluate two tasks: window classification and object detection. 

\vspace{-3mm}
\subsubsection*{Window classification} 
\vspace{-2mm}
For this experiment, we use all positive windows ($IoU \geq 0.5$) in the test set \texttt{val$_2$} and 1 million randomly sampled negative ones ($IoU < 0.5$).
We evaluate window classification in terms of two measures: false positives at test recall, and average precision. 

%\vspace{-2mm}
\noindent \textit{False positives at test recall.}
This measure counts how many false positives are produced on the test set, at the recall point produced by the ensemble of E-SVMs calibrated by our method. Note this is exactly what our calibration procedure optimizes for.
Given the thresholds $\mathbf{\Theta}$ output by our algorithm (sec.~\ref{sec:algo}), we compute recall as the percentage of positive windows scored positively by the ensemble on the test set (top row of table~\ref{table:exp1}). Interestingly, the thresholds generalize well to test data and lead to high recall on almost all classes. 

%\vspace{-1mm}
We compare several methods at this recall point. The main four are EE-SVM with no calibration, EE-SVM with independent sigmoid calibration~\cite{MalisiewiczICCV11}, our joint calibration fitting thresholds and our joint calibration with sigmoid (table~\ref{table:exp1}, rows 2-5).
As expected, EE-SVM with no calibration performs very poorly and some form of calibration is necessary.
Our joint calibration method considerably outperforms independent calibration,
and the version with sigmoid brings another small boost in performance.
These results demonstrate the benefit of our joint calibration, that takes into account the max operation of the EE-SVM. Given these results, we omit EE-SVM with no calibration from further analysis.
Table~\ref{table:exp1} also presents results for isotonic regression, affine calibration and R-CNN.
On average, our joint calibration outperforms all these methods, albeit by a smaller margin. 

%\vspace{-1mm}
\noindent \textit{Average precision.}
In table~\ref{table:exp2} we compare techniques in terms of average precision. As this measure requires a continuous score of test windows, we only consider our joint calibration with sigmoid. 
Joint calibration outperforms independent sigmoid calibration on all classes, and improves mAP by 2.3\%.
This further highlights the benefits of joint calibration, in a scenario that is not exactly what it was optimized for.
Table~\ref{table:exp2} also presents results for isotonic regression, affine calibration and R-CNN. On average, joint calibration performs better than all these methods, abeit by a modest margin (about +1\% mAP).

%\vspace{-3mm}
\subsubsection*{Object detection} 
\vspace{-3mm}
We evaluate our joint calibration method against independent sigmoid calibration on the task of object detection.
Note how this task adds a layer of non-maximum suppression (NMS) to the pipeline.
As our calibration procedure does not take into account NMS, it is not obvious that the benefit seen so far on window classification will carry over to object detection.
As table~\ref{table:exp3} shows, joint calibration outperforms independent sigmoid calibration on this task as well (+3.3\% mAP). Joint calibration performs equally or better on all classes but \textit{koala}. For some classes the improvement is substantial: +14\% AP on \textit{baseball} and +14.7\% AP on \textit{butterfly}. 

Furthermore, table ~\ref{table:exp3} also presents results for isotonic regression, affine calibration and R-CNN. Isotonic regression performs comparably to independent sigmoid calibration, whereas affine calibration delivers about the same mAP as our joint calibration.
Interestingly, R-CNN does considerably better than all other methods, including our EE-SVM with joint calibration.
This is somewhat surprising, as EE-SVM was shown much better than a single linear SVM on HOG features~\cite{MalisiewiczICCV11}. We attribute this phenomenon to the CNN features, which are more easily linearly separable \cite{donahue13decaf,girshick14cvpr,razavian14cvpr, chatfield14bmvc}.
Besides, note that despite the high performance, R-CNN lacks the crucial ability of EE-SVM to associate training exemplars to objects detected in test images, and it is therefore not suitable for annotation transfer. 

%showed that HOG features are not that separable and EE-SVM double the performance of a category-wise linear SVM trained on HOG features. 

% ------------------------------------------------------------------------------------ APPROXIMATE
\subsection{Approximate joint calibration}
\label{sec:exp_approx}
\vspace{-2mm}
In this section we evaluate our approximate joint calibration technique (sec.~\ref{sec:approx}). By relaxing global optimality, we can find a feasible solution even for large problems.

\vspace{-3mm}
\paragraph{ILSVRC2014.}
We experiment here by training and calibrating EE-SVM on the union of \texttt{val$_1$} and \texttt{train}. This results in a large number of E-SVMs. Each class has between 640 and 2000 E-SVMs, and between 3000 and 13000 positive windows $\mathcal{P}$.
%It is clear from these numbers that the search tree is massive. 
%
We report results on the task of window classification averaged over the 10 classes.
Note that these results cannot be compared to the ones in tables~\ref{table:exp1} and ~\ref{table:exp2} because here we have larger training and test sets.
% \todo{DVD: how can we motivate the fact that we don't have results for detection?}. \\

\noindent \textit{False positives at test recall.}
As table~\ref{table:ap_exp1} shows, our approximate joint calibration procedure still achieves high recall while returning much fewer false positives than no calibration and independent sigmoid calibration.
When adding a sigmoid our calibration improves even further by a good margin. This shows that our method provides an excellent association between positive windows and E-SVMs.

\noindent \textit{Average precision.} Results are presented in table~\ref{table:ap_exp2}. Joint calibration improves over independent sigmoid calibration by +3.2\% mAP.

\begin{table}
\small \centering
{
\begin{tabular}{ | l | c | }
\hline
{\bf ILSVRC 2014 - trained on \texttt{Val$\mathbf{_1}$+train }} & mean\\\hline
\it  Recall & 85.2 \\\hdashline
%& \it (899/955) & \it (1795/1992) & \it (760/776) & \it (2187/2349) & \it (3584/3675) & \it (623/645) & \it (660/830) & \it (669/756) & \it (1311/1326) & \it (424/497) \\\hdashline
EE-SVM no calibration & 137k\\
EE-SVM indep. sigmoid calibration \cite{MalisiewiczICCV11} & 107k\\
EE-SVM joint calibration & 82k \\
EE-SVM joint calibration w/ sigmoid & \bf 54k \\\hline
\end{tabular}}
\caption{\small \it \textbf{Window classification - False positives at test recall.} Results on a subset of ILSVRC 2014 \texttt{Val$_2$} (mean over 10 classes). We used all positive windows and 2 million randomly sampled negative ones. \vspace{-1mm}}
\vspace{-1mm}
\label{table:ap_exp1}
\end{table}

\begin{table}
\small \centering
{
\begin{tabular}{ | l | c | }
\hline
{\bf ILSVRC 2014 - trained on \texttt{Val$\mathbf{_1}$+train }} & mAP\\\hline
EE-SVM indep. sigmoid calibration \cite{MalisiewiczICCV11} &  39.7\\
EE-SVM joint calibration w/ sigmoid & \bf 42.9 \\\hline
\end{tabular}}
\caption{\small \it \textbf{Window classification - Average precision.} Results on a subset of ILSVRC 2014 \texttt{Val$_2$} (mean over 10 classes, same data as table~\ref{table:ap_exp1}). \vspace{-3mm}}
\vspace{-2mm}
 \label{table:ap_exp2}
\end{table}

%-------------------------------------------------------------------------
\vspace{-4mm}
\paragraph{PASCAL VOC 2007.}
In order to compare against the original EE-SVM of \cite{MalisiewiczICCV11}, we experiment here on the PASCAL VOC 2007 dataset. We train and calibrate EE-SVM on the \texttt{trainval} subset and evaluate them on \texttt{test}.
We report results on object detection in terms of mAP over the 20 classes (table~\ref{table:pascal}).
We compare traditional EE-SVM on HOG features (19.8 mAP, as reported by \cite{MalisiewiczICCV11}),
independently calibrated EE-SVM on CNNs (40.8 mAP), and our joint calibration on the same features (42.7 mAP).
These results highlight two points:
(1) joint calibration improves over independent calibration by +2\% mAP, confirming what observed on ILSVRC2014;
(2) CNN features bring a huge improvement over HOG to EE-SVM models (doubling mAP in this case).
This confirms recent findings \cite{girshick14cvpr} about the benefits of CNN features for object detection.

% ------------------------------------------------------------------------------------ PRUNING
\subsection{Pruning statistics and runtimes} \label{sec:pruning}
\vspace{-1mm}

\paragraph{Pruning statistics.}
Here we experimentally evaluate how effective our pruning techniques of sec.~\ref{sec:efficient_search} are.
Observation 3 (\textit{\small line 2}, Algo.~\ref{algo}) reduces the depth of the tree by 20\%, on average. 
Observation 4 (\textit{\small line 3}) improves pruning by bound immensely. In a small problem with 100 E-SVMs we tried ordering the positive windows $\mathcal{P}$ randomly. The algorithm took
%(after feature extraction and some data pre-computation)
two hours to find the global solution. On the other hand, when sorting $\mathcal{P}$ according to observation 4, the algorithm found the same solution in about 2 minutes. 
Observations 1 and 2 also bring a substantial speed-up. After finding a first feasible solution (\textit{\small line 1}), the algorithm (\textit{\small lines 8,9}) prunes 40\% of the nodes it visits on problems with 100 E-SVMs, and 70\% on problems with 1000 E-SVMs. 
%Finally, we also note that the configuration of thresholds automatically discards some E-SVMs. These are the ones that started with a very high threshold (as $e_2$ with $\theta_2^1$ in fig. \ref{fig:pipeline}) and to which the algorithm never associated any positive window. On average, the algorithm discards 25\% of the E-SVMs.

\vspace{-5mm}
\paragraph{Runtime.}
We measure runtimes on a 4-cores Intel Core i5 2.0GHz.
Exhaustive search is extremely inefficient and takes 15h to find the globally optimal solution for a tiny problem with 15 E-SVMs and 50 positive windows. Our efficient and exact algorithm (sec.~\ref{sec:efficient_search}) finds the same solution in just a few seconds.
%In our implementation we pre-compute the false positives produced by every possible candidate threshold. This reduces the computation of the loss function $\mathcal{L}(\mathbf{\Theta})$ during the search to a single \texttt{unique} operation. On this takes roughly 10 minutes for 500 E-SVMs, 5k positive and 250k negative windows. 
This algorithm scales up to problems with about 200 E-SVMs and 4k positive windows in reasonable time (a few hours).
For larger problems we rely on our approximate search algorithm (sec.~\ref{sec:approx}).
While we let it run for several hours, in most cases the loss stops decreasing significantly already after a few minutes.

\begin{table}
\small \centering
%\resizebox{\columnwidth}{!}
{
\begin{tabular}{ | c | c | c | c | } 
\hline
{\bf PASCAL VOC}  & \multirow{2}{*}{Feature} & \multirow{2}{*}{Calibration} & \multirow{2}{*}{mAP}\\
{\bf 2007 test} & & & \\\hline
\multirow{3}{*}{EE-SVM} & HOG & independent & 19.8 \\
& CNN & independent &  40.8\\
& CNN & joint & 42.7 \\\hline
\end{tabular}}
\caption{\small \it \textbf{Object detection - Average precision.} Results on PASCAL VOC 2007 \texttt{test} (mean over 20 classes). EE-SVM HOG results are from \cite{MalisiewiczICCV11}. \vspace{-2mm}}
\vspace{-2mm}
\label{table:pascal}
\end{table}

% ------------------------------------------------------------------------------------ CONCLUSIONS
%\vspace{-1mm}
\section{Conclusion} \label{sec:concl}
\vspace{-1mm}
We presented a method for calibrating the Ensemble of Exemplar SVMs model. While the standard approach calibrates each SVM independently, our method optimizes their joint performance as an ensemble. We formulated joint calibration as a constrained optimization problem and devised an efficient optimization algorithm to find its global optimum.
In order to make the optimization computationally feasible, the algorithm dynamically discards parts of the solution space that cannot contain the optimum, by exploiting four observations about the structure of the problem.

We presented experiments on 10 classes from the ILSVRC 2014 dataset and 20 from PASCAL VOC 2007.
Our joint calibration procedure outperforms the classic independent sigmoid calibration by a considerable margin on the task of classifying windows as belonging to an object class or not. On object detection, this better window classifier leads to an improvement of about 3\% mAP.

%\clearpage
\paragraph*{Acknowledgement.}
We gratefully acknowledges support by SNSF fellowship PBEZP-2142889 and ERC Starting Grant VisCul.

%\vspace{-3.5mm}
{
\bibliographystyle{ieeetr}
\bibliography{../../bibtex/shortstrings,../../bibtex/vggroup,../../bibtex/calvin}
}
\end{document}